\documentclass[11pt,a4paper]{article}
\usepackage[hyperref]{emnlp-ijcnlp-2019}
\usepackage{times}
\usepackage{latexsym}
\usepackage{xcolor}
\usepackage{graphicx}
\usepackage{mathtools, nccmath}
\usepackage{amsfonts}
\usepackage{url}
\usepackage{dblfloatfix}
\usepackage[T1]{fontenc}
\usepackage[polish]{babel}
\usepackage[utf8]{inputenc}
\usepackage{longtable}
\usepackage{tipa}

\aclfinalcopy 

\title{TMLab: Generative Enhanced Model (GEM) for adversarial attacks} 

\author{Piotr Niewiński, Maria Pszona, Maria Janicka\\
 Samsung R\&D Institute Poland \\
 {\tt \{p.niewinski, m.pszona, m.janicka\}@samsung.com} \\
}

\date{}

\begin{document}
\maketitle
\begin{abstract}
We present our Generative Enhanced Model (GEM) that we used to create samples awarded the first prize on the FEVER 2.0 \textit{Breakers} Task. GEM is the extended language model developed upon GPT-2 architecture. The addition of novel target vocabulary input to the already existing context input enabled controlled text generation.
The training procedure resulted in creating a model that inherited the knowledge of pretrained GPT-2, and therefore was ready to generate natural-like English sentences in the task domain with some additional control.
As a result, GEM generated malicious claims that mixed facts from various articles, so it became difficult to classify their truthfulness.
\end{abstract}

\section{Introduction}
Fact-checking systems usually consist of separate modules devoted to information retrieval (IR) and recognizing textual entailment (RTE), also known as natural language inference (NLI). First, information retrieval module searches through the database in order to find sentences related to the given statement. Next, entailment module, with respect to the extracted sentences, classifies the given claim as TRUE, FALSE or NOT ENOUGH INFO. Currently, the best results are achieved by pretrained language models that are fine-tuned with task specific data \cite{Yang2019, Liu2019}.

Our task was to provide adversarial examples to break fact-checking systems.
Since many fact-checking systems are based on neural language models, they might be less resistant to attacks with samples prepared within the same approach. In line with recent advances in natural language generation, we used the GPT-2 model \cite{Radford2019}, which we modified to prepare malicious adversarial examples. GPT-2 generates subsequent sentences based on a given textual context and originally was trained on the WebText corpus. Our GEM architecture was expanded with \textit{target} input for controlled generation, and carefully trained on the task data.
During inference, the model was fed with the Wikipedia content. Simultaneously, \textit{target} input was provided with named entities, terms and phrases extracted from Wikipedia articles. 

\section{FEVER \textit{Breakers} Subtask}
The second edition of Fact Extraction and Verification (FEVER 2.0) shared task was the three-phased contest utilizing the idea of adversarial training \cite{Thorne2019}. In the first phase, \textit{Builders} had to create a fact-checking system. This system should extract evidence sentences for a given claim from Wikipedia articles that either SUPPORT or REFUTE this claim. It can also classify an example as NOT ENOUGH INFO. In the second phase, \textit{Breakers} had to supply malicious examples to fool the existing systems. Finally, \textit{Fixers} were obliged to improve those systems to withstand adversarial attacks.
The model presented in this paper originated as a part of \textit{Breakers} subtask. The aim of this task was to create adversarial examples that will break the majority of systems created in the \textit{Builders} phase. Malicious claims could have been generated automatically or manually and were supposed to be balanced over three categories. The evidence sentences had to be provided in the SUPPORTS and REFUTES categories. 

\section{Generative Enhanced Model}
\begin{table*}[!b]
\centering
\begin{tabular}{lp{14cm}}
\hline
\hline
\textit{context} & The main railway stations of the province are Bydgoszcz and Toruń. Both stations are served by fast PKP Intercity trains which connect them with the capital Warsaw, as well as other major Polish cities.\\
\textit{target} & \textbf{characterized Portugal farmland numerous lakes and forests}\\
\textit{output} & \leavevmode\color{blue} Bydgoszcz is characterized as a medium-sized city in Portugal, with its farmland and numerous lakes and forests.\\
\hline
\textit{context} & Near the beginning of his career, Einstein thought that Newtonian mechanics was no longer enough to reconcile the laws of classical mechanics with the laws of the electromagnetic field. This led him to develop his special theory of relativity during his time at the Swiss Patent Office in Bern (1902--1909).\\
\textit{target} & \textbf{objected quantum mechanics contrast Bohr}\\
\textit{output} &  \leavevmode\color{blue} Einstein objected to the use of quantum mechanics in contrast with Bohr's theory of gravitation, which he thought was the most superior theory of relativity.\\
\hline
\textit{context} & The City of New York, usually called either New York City (NYC) or simply New York (NY), is the most populous city in the United States. With an estimated 2018 population of 8,398,748 distributed over a land area of about 302.6 square miles (784 km2), New York is also the most densely populated major city in the United States.\\
\textit{target} & \textbf{realized asset establishment independent border}\\
\textit{output} & \leavevmode\color{blue} New York City is realized as an economic, cultural, and political asset upon the establishment of an independent border country.\\
\hline
\textit{context} & Lasse Hoile (born 1973 in Aarhus, Denmark) is an artist, photographer and film-maker. He has collaborated with musician Steven Wilson and his projects Porcupine Tree and Blackfield. He has also designed live visuals for the US progressive metal band Dream Theater.\\
\textit{target} & \textbf{true fact Swedish progressive metal band Stockholm}\\
\textit{output} & \leavevmode\color{blue} Hoile's true interest is in fact the Swedish progressive metal band, Stockholm.\\
\hline
\hline
\end{tabular}
\caption{Examples of first sentences generated for given context and target words.}
\label{tab:sample}
\end{table*}

\subsection{Natural Language Generation with Neural Networks}
Neural language models, such as GPT-2, rely on modeling conditional probability of an oncoming token for a given input sequence (context). 
Given the dictionary of tokens $\mathbb{D}$ and sequence $x_0\ldots x_N\, (x_i \in \mathbb{D})$ model computes conditional probability for every token $x$ from $\mathbb{D}$:
$$
p(x) = \prod_{i=1}^{n}p(x_n|x_1, \ldots, x_{n-1}).
$$
During each stage of the process, the language model outputs probability distribution of tokens from dictionary $\mathbb{D}$. There are various approaches to select a single token from output distribution. Usually, the one with the highest probability is chosen or is sampled from the distribution. This distribution may be slightly modified by parameters like temperature and top-$k$. However, such context-based language generation gives us very little, if any, control over the model output.

Taking that into account, our main goal was to modify the architecture of Generative Pretraining Transformer (GPT-2), and enable additional control during the generation process. Therefore, GEM samples subsequent tokens by using information from two inputs: \textit{context} (past) and \textit{target}. As target words various combinations of English nouns, verbs, and named entities can be provided and their number may vary.

GEM stops generating output when the total number of consecutive tokens reaches the value of parameter \textit{maxTokens}. As a consequence, not only the first sentence, for which target words are given, is generated. That kind of generation procedure is expected to keep the original model's ability to build sentences even without target words. The examples of first sentences from the model output are presented in Table \ref{tab:sample}.

\subsection{Architecture}

\begin{figure*}[!t]
\begin{center}
\includegraphics[width=1.0\textwidth]{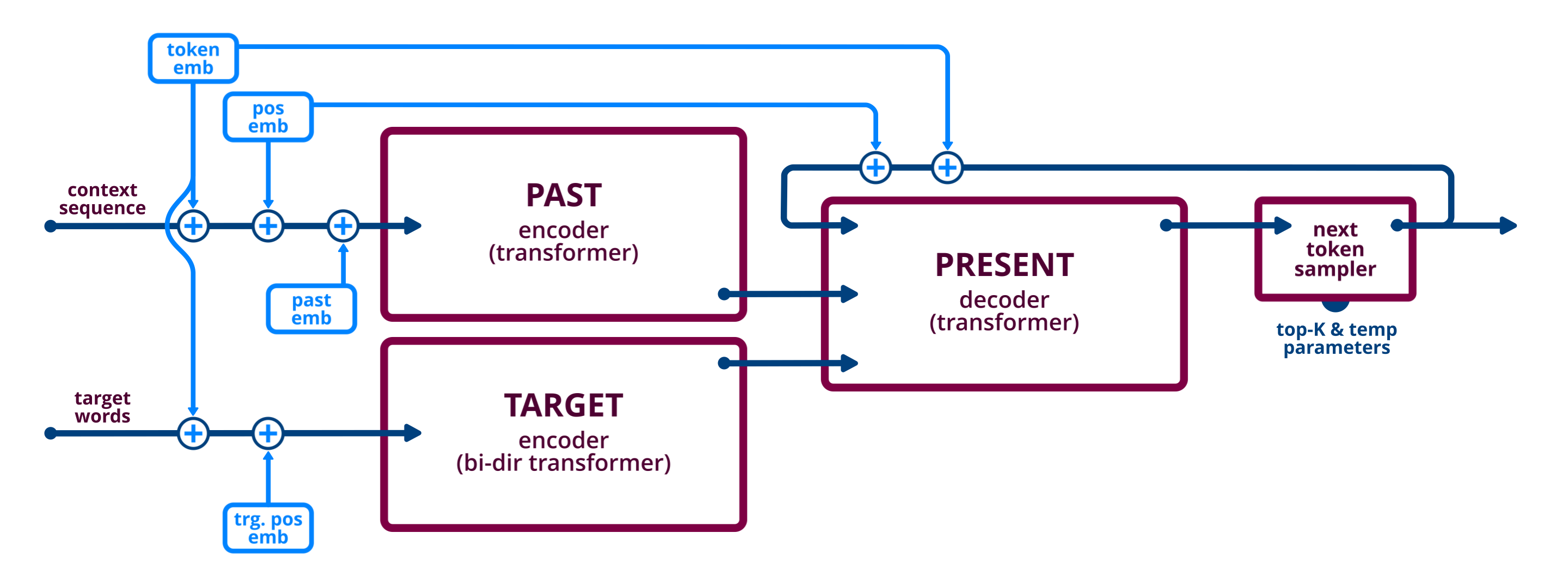}
\caption{The architecture of GEM.}
\label{fig:GEM:architecture}
\end{center}
\end{figure*}

GEM is build upon encoder-decoder Transformer-based language model architecture \cite{Vaswani2017} enhanced with second Transformer encoder for \textit{target} words.

Typical autoregressive neural language model, such as GPT-2, generates the next token using representations of \textit{context} tokens (past) and \textit{present} tokens (previously generated). Given \textit{context} tokens $c_1\ldots c_n$ and \textit{present} tokens $p_1\ldots p_n$, model encodes \textit{context} representations $cr_1\ldots cr_n$ and \textit{present} representations $pr_1\ldots pr_n$. With concatenated representations of \textit{context} and \textit{present} $cr_1\ldots cr_n ; pr1\ldots pr_n$ model generates the next token $p_{n+1}$.

Both \textit{context} and \textit{present} representations are prepared with Transformers, using same shared parameters and embeddings. Such concept minimizes the number of parameters, and is optimal for classic generation task. Representations of \textit{context} and \textit{present} when concatenated are undifferentiated for decoder attention mechanism - the decoder has no information where \textit{context} ends and \textit{present} starts. This is not a problem for a standard task of neural language modeling.

\begin{figure*}[t!]
\begin{center}
\includegraphics[width=1.03\textwidth]{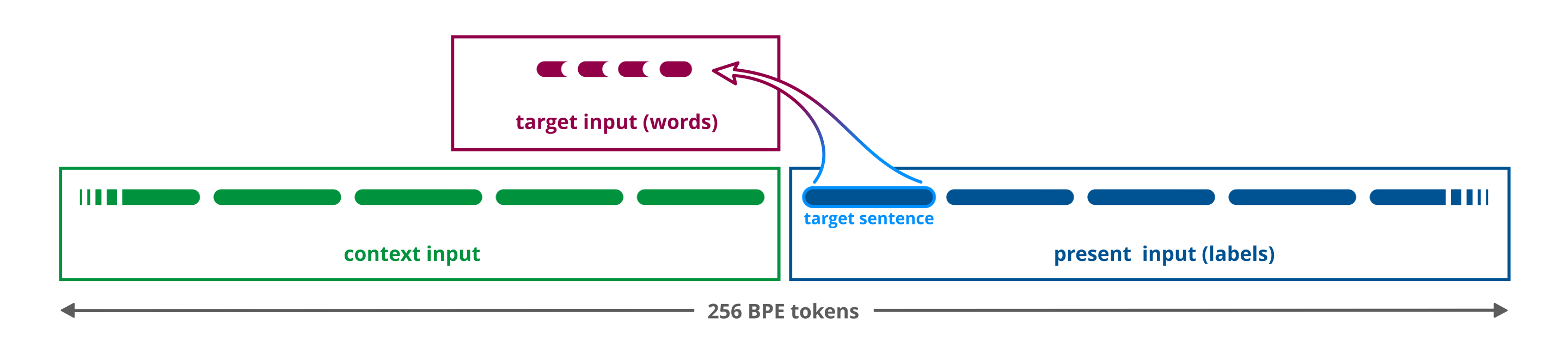}
\caption{GEM training sample.}
\label{fig:GEM:sample}
\end{center}
\end{figure*}

The GEM's architecture is outlined in Figure \ref{fig:GEM:architecture}. Just like GPT-2, the proposed model uses concatenated representations. However, in GEM \textit{target} words representations  $tr_1\ldots tr_n$, prepared by target encoder, are added:
$$
cr_1\ldots cr_n ; tr_1\ldots tr_n ; pr_1\ldots pr_n.
$$
In contrast to standard neural language models, GEM, in order to work properly, needs to differentiate between all three sources of representations. Both positional embeddings and Transformer weights of target encoder are not shared with past encoder and present decoder, and are initialized from scratch instead (with random normal initializer of 0.02 standard deviation). During the training, GEM learns the weights of target encoder to properly accomplish the task and distinguish \textit{target} representations from the other two: \textit{context} and \textit{present}. In order to pass the information about the origin of \textit{context} representations, we have added \textit{past} embedding (single trainable vector) to \textit{context} tokens.

Past encoder and present decoder were initialized with GPT-2 checkpoint parameters. The idea was to use the knowledge of pretrained state-of-the-art English language model. Token embeddings from GPT-2 checkpoint were not updated while training. The final size of GEM is equal to 170-190\% of the original GPT-2 model size (depending on GPT-2 version).

\subsection{Training Procedure}

The model was trained on the corpus provided by FEVER organizers. It contains a dump taken from the English-language version of Wikipedia from 2017. Each article was sentence-tokenized with spaCy tokenizer \cite{spacy2}, and then each sentence was tokenized with BPE tokens from GPT-2 model.

Single training sample was prepared with the following procedure. First, random \textit{target} sentence from the given Wikipedia articles was chosen. Next, the arbitrary number of words ranging from 20\% to 60\% was selected from \textit{target} sentence. Selected words built \textit{target} input. In  addition, a small number of random words (up to 10\%), which do not appear in \textit{target} sentence, may be added to the set of \textit{target} words. The intuition behind adding the noise to \textit{target} words during the training phase was that it would prevent the model from directly 'copying' from \textit{target} input. The model was supposed to decide whether to include the given words or not, because some of them may be irrelevant. Sentences forerunning \textit{target} sentence established \textit{context} input. As a result, \textit{target} sentence with the following sentences served as gold labels. In addition, single training sample was limited to 256 BPE tokens, which on average corresponds to 10 sentences. Single training sample is presented in Figure \ref{fig:GEM:sample}.

We have fine-tuned the original GPT-2 language model with the text generation task on the FEVER Wikipedia data (30M sentences). The model was fed with the Wikipedia content, and was asked to generate next sentences.
GPT-2, without additional training, managed to achieve 37\% accuracy on the stated task. It means that 37\% of tokens generated by the model matched the gold labels from the original Wikipedia text. However, not modified GPT-2 fine-tuned with the given Wikipedia data was able to achieve 43\% accuracy on a validation set.

\begin{table*}[b!]
\centering
\begin{tabular}{lp{14cm}}
\hline
\hline
\textit{wiki-A} & Joseph Cao\\
\textit{wiki-B} & Republican Party (United States)\\
\textit{context} & Ánh Quang ''Joseph'' Cao (\textipa{[ 'gaU]}; Cao Quang Ánh born March 13, 1967) is a Vietnamese American politician who was the U.S. Representative for from 2009 to 2011. In April 2011, Cao announced his candidacy for the office of Attorney General of Louisiana; however, in September 2011 he pulled out of the race, and the incumbent Buddy Caldwell ran unopposed for a second term. He is the first Vietnamese American to serve in Congress, and the first and thus far only Republican from his New Orleans-based district since 1891. In December 2015, he announced that he would run for the open U.S. Senate seat being vacated by retiring fellow Republican David Vitter in 2016. The Republican Party, commonly referred to as the GOP (abbreviation for Grand Old Party), is one of the two major contemporary political parties in the United States, the other being its historic rival, the Democratic Party. He is a member of the Republican Party.\\
\textit{target} & \textbf{The party named dominant value during}\\
\textit{output} & \leavevmode\color{blue} Joseph Cao was elected to Congress in 2009 and has named a number of prominent Republicans to be the dominant value players during his time in the House.\\
\hline
\textit{wiki-A} & Remmina\\
\textit{wiki-B} & Remote Desktop Protocol\\
\textit{context} & Remmina is a remote desktop software client for POSIX-based computer operating systems. Remmina is in the package repositories for Debian versions 6 (Squeeze) and later and for Ubuntu versions since 10.04 (Lucid Lynx). As of 11.04 (Natty Narwhal), it replaced tsclient as Ubuntu's default remote desktop client. The FreeBSD ports/package collection also contains it as a separate port and additional protocol-specific plugin ports. Remote Desktop Protocol (RDP) is a proprietary protocol developed by Microsoft, which provides a user with a graphical interface to connect to another computer over a network connection. It supports the RDP, VNC, NX, XDMCP, SPICE and SSH protocols.\\
\textit{target} & \textbf{RDP client for must run software} \\
\textit{output} & \leavevmode\color{blue} Remmina is a standalone RDP client for Windows and must run as a user on a Linux system, or the client software will be unavailable.\\
\hline
\hline
\end{tabular}
\caption{Examples of generated sentences for given context and target words.}
\label{tab:sample2}
\end{table*}

Naturally, we expected higher accuracy with additional target words input. Though, we were afraid that adding new parameters and modifying the architecture might result in a significant loss of GPT-2 pretrained knowledge. During the training process, the initial accuracy of GEM was 3\% and it raised very quickly. After the first epoch of training it achieved 47\%. We trained the model with batches of 16 samples for 6 epochs, and the learning rate was set to 1e-5. The batch size of training data was limited by the memory of GPU, while other hyperparameters were chosen with the grid search evaluation.
As a result, GEM finally achieved 53\% accuracy while still not overfitting the data. High final accuracy of GEM states that the knowledge of GPT-2 was not forgotten, and, at the same time, the model learned to effectively use the provided target words.

We can estimate the theoretical maximum accuracy (higher bound) of GEM with stated task and training scheme. Each training sample, on average, corresponds to 10 sentences. The model generates tokens for 5 sentences. GEM additionally fed with target words is able to achieve the maximum accuracy of 100\% for the first sentence, and keep the maximum accuracy of fine-tuned GPT-2 (43\%) for the remaining four sentences. With these assumptions, the average accuracy across the entire sample would reach 54.4\%. Therefore, our final 53\% accuracy is only a bit lower, and reversing these calculations we can get up to 93\% accuracy for the first sentence when GEM is supported with target input.

\section{Claims generation procedure}

The procedure of generating claims was driven by the assumption that sophisticated claims contain knowledge from many sources, and cannot be checked with a single evidence sentence. To force automatic generation of such claims, we have built pipeline for input data preparation and claims selection described below.
Wikipedia articles have a hypertext form with references to other articles. A single input sample (context and target words) was based on two Wikipedia articles: wiki-A and wiki-B. Wiki-A was randomly selected from the corpus. A set $\mathbb{B}$ was created from articles hyperlinked in the first five sentences of wiki-A. Then, it was filtered with the following principles. 
An article $b$ was removed from $\mathbb{B}$ if:
\begin{itemize}
\item any words from title of $b$ appeared in wiki-A title
\item $b$ hyperlink (string) in wiki-A was equal to $b$ title
\end{itemize}
Finally, wiki-B article was randomly selected from $\mathbb{B}$. 

The target words were randomly selected from the second sentence of wiki-B. Similar to the training procedure, their number varied from 20\% to 60\% of source-sentence words. Context sentences were composed of mixed wiki-A and wiki-B sentences, excluding sentences containing hyperlinks to wiki-B and the second sentence of wiki-B. Finally, the title of wiki-A article was appended to the context. GEM started generation from this point.

Generated claims were further filtered, and the ones meeting any of the listed conditions were removed:
\begin{itemize}
\item claims not ending with a dot (probably due to incorrect tokenization)
\item claims shorter than 30 characters and longer than 200
\item claims containing \texttt{<endoftext>} token
\item claims too similar to the first sentence of wiki-A (measured with \citet{levenshtein} distance)
\item claims containing numbers and dates not appearing in wiki-A article
\item claims containing any words out-of-vocabulary, where vocabulary was built from words of all Wikipedia articles
\end{itemize}
The examples of generated claims are shown in Table \ref{tab:sample2}. 

\begin{figure*}[!t]
\begin{center}
\includegraphics[width=1.0\textwidth]{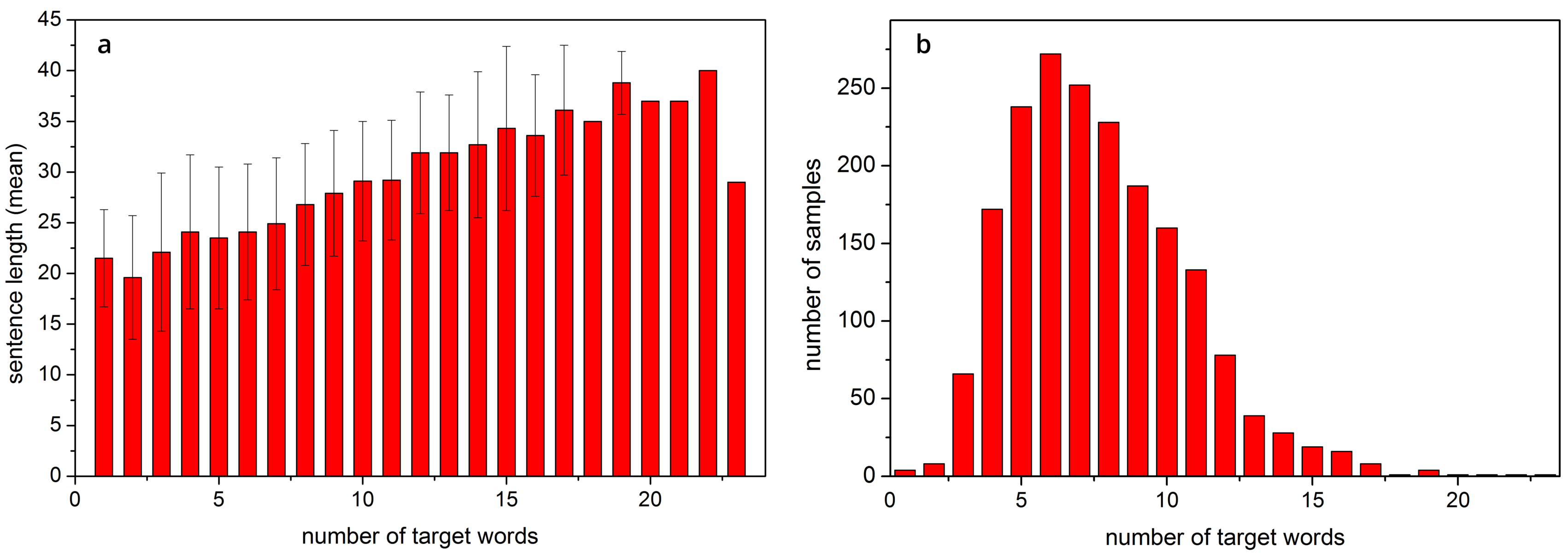}
\caption{The dependency of the length of generated sentences on the number of target words (a) and target words statistics (b).}
\label{fig:statistics}
\end{center}
\end{figure*}

\begin{table*}[b!]
\centering
\begin{tabular}{lp{10.5cm}}
\hline
\hline
\textit{double negation} & It is not true that one can falsely say that double negation theorem states that "If a statement is false, than it is not the case that the statement is not false."\\
\hline
\textit{comparison} & Łączka does not lay as close to Siedlce as Żuków.\\
\hline
\textit{paraphrase} &  Finding a theory of everything, which is considered a final theory,  still remains a challenge.\\
\hline
\textit{polysemy} & There is a fashion house with a word meaning 'sweet' in its name.\\
\hline
\textit{negation} & K2 is not the highest mountain in the world.\\
\hline
\hline
\end{tabular}
\caption{Examples of manually prepared samples.}
\label{tab:samples:manual}
\end{table*}

The dependency between the number of provided target words and the length of generated sentence is presented in Figure \ref{fig:statistics}a. The statistics of target words number is shown in Figure \ref{fig:statistics}b. The presented results are based on 1917 samples generated by GEM model and clearly indicates the correlation between the length of generated sentence and number of the target words: the fewer words the system gets, the shorter sentence will be generated.

The automatically generated claims required further manual labeling as SUPPORTS, REFUTES or NOT ENOUGH INFO. Moreover, in the case of the first two classes, the evidence sentences from Wikipedia were supposed to be delivered. 
Initially, each claim was annotated independently by two linguists. Both annotators agreed on 58.5\% samples. The distribution of labels was highly unbalanced: 72.6\% REFUTES, 13.7\% NOT ENOUGH INFO, and 4.3\% SUPPORTS. The remaining 9.4\% of samples contained language errors. Finally, the supporting sentences from Wikipedia were manually extracted. 

Due to a small number of claims labeled SUPPORTS in automatically generated data, there was a need to manually create some examples in this category. Malicious claims were based on several tricks, such as the usage of double negation, polysemy, comparison (of age, area, population), calculations, paraphrase (e.g. using phrases from Wikipedia articles unrelated to a claim or evidence), complex chains of reasoning, etc. The examples are provided in Table \ref{tab:samples:manual}.

\section{Results}
Our adversarial attack was ranked the first place in the official FEVER 2.0 results.
In total, we submitted 155 various claims (104 automatically generated and 51 written by human), which were divided into train and test sets. The quality of the test set was described by three measures: Correct Rate, Raw Potency and Potency, all defined in \citet{Thorne2019}. The Correct Rate, which is a percentage of positively verified samples, was 84.81\%. This means that the organizers disqualified about 15\% of our claims, mostly due to grammatical errors, such as word repetitions or wrong verb forms.   
The Raw Potency of the prepared adversarial examples, defined as the percentage of incorrect predictions, averaged over all systems was 78.80\%.
Finally, the main evaluation measure - Potency (the Raw Potency scaled by the Correct Rate) achieved by our samples was 66.83\%.

\section{Conclusions}
The claims provided by GEM model appeared to be the most challenging for fact-checking systems competing in a FEVER 2.0 shared task. Our strategy was to mix Wikipedia articles, which were connected to each other with a hyperlink and filtered with the established strategies. This approach led to generating cohesive, well-structured samples, which were challenging for automated verification. As GEM was developed upon GPT-2 architecture, and inherited its knowledge, the model might be biased towards factual inaccuracies. The established pipeline could just strengthen this tendency, which finally reflected in the class imbalance of automatically generated content. Automatic generation of complex claims supported by Wikipedia would require fine-tuned procedures. This issue seems to be an interesting challenge that could be addressed in further research.

The preparation of adversarial examples is a very prominent concept of modern machine learning research area. It gives the possibility of fast, automated, and massive generation of additional samples. Importantly, injecting the malicious examples into training data may result in more robust and accurate models. GEM designed for controlled text generation can also be applied in various text-driven systems, e.g. conversational agents, text summarizers or style transfer models.

\bibliography{emnlp-ijcnlp-2019}

\begin{thebibliography}{7}
\expandafter\ifx\csname natexlab\endcsname\relax\def\natexlab#1{#1}\fi

\bibitem[{Honnibal and Montani(2017)}]{spacy2}
Matthew Honnibal and Ines Montani. 2017.
\newblock {spaCy 2}: Natural language understanding with {B}loom embeddings,
  convolutional neural networks and incremental parsing.
\newblock To appear.

\bibitem[{Levenshtein(1966)}]{levenshtein}
VI~Levenshtein. 1966.
\newblock {Binary Codes Capable of Correcting Deletions, Insertions and
  Reversals}.
\newblock \emph{Soviet Physics Doklady}, 10:707.

\bibitem[{Liu et~al.(2019)Liu, Ott, Goyal, Du, Joshi, Chen, Levy, Lewis,
  Zettlemoyer, and Stoyanov}]{Liu2019}
Yinhan Liu, Myle Ott, Naman Goyal, Jingfei Du, Mandar Joshi, Danqi Chen, Omer
  Levy, Mike Lewis, Luke Zettlemoyer, and Veselin Stoyanov. 2019.
\newblock \href {http://arxiv.org/abs/1907.11692} {Roberta: {A} robustly
  optimized {BERT} pretraining approach}.
\newblock \emph{CoRR}, abs/1907.11692.

\bibitem[{Radford et~al.(2019)Radford, Wu, Child, Luan, Amodei, and
  Sutskever}]{Radford2019}
Alec Radford, Jeff Wu, Rewon Child, David Luan, Dario Amodei, and Ilya
  Sutskever. 2019.
\newblock Language models are unsupervised multitask learners.

\bibitem[{Thorne and Vlachos(2019)}]{Thorne2019}
James Thorne and Andreas Vlachos. 2019.
\newblock \href {http://arxiv.org/abs/1903.05543} {Adversarial attacks against
  fact extraction and verification}.
\newblock \emph{CoRR}, abs/1903.05543.

\bibitem[{Vaswani et~al.(2017)Vaswani, Shazeer, Parmar, Uszkoreit, Jones,
  Gomez, Kaiser, and Polosukhin}]{Vaswani2017}
Ashish Vaswani, Noam Shazeer, Niki Parmar, Jakob Uszkoreit, Llion Jones,
  Aidan~N. Gomez, Lukasz Kaiser, and Illia Polosukhin. 2017.
\newblock \href {http://arxiv.org/abs/1706.03762} {Attention is all you need}.
\newblock \emph{CoRR}, abs/1706.03762.

\bibitem[{Yang et~al.(2019)Yang, Dai, Yang, Carbonell, Salakhutdinov, and
  Le}]{Yang2019}
Zhilin Yang, Zihang Dai, Yiming Yang, Jaime~G. Carbonell, Ruslan Salakhutdinov,
  and Quoc~V. Le. 2019.
\newblock \href {http://arxiv.org/abs/1906.08237} {Xlnet: Generalized
  autoregressive pretraining for language understanding}.
\newblock \emph{CoRR}, abs/1906.08237.

\end{thebibliography}
\bibliographystyle{acl_natbib}

\appendix


\end{document}


\appendix
\begin{longtable}{lp{13cm}} \multicolumn{2}{r}{{Continued on next page}} \\ 
\endfoot
\multicolumn{2}{r}{} \\ 
\endlastfoot
\hline
\hline
\textit{wiki-A} & Andrey Strokin\\
\textit{wiki-B} & 2000 Summer Paralympics\\
\textit{context} & Andrey Strokin is a paralympic swimmer from Russia competing mainly in category S13 events. In 2000 he finished in fourth in the 400 m freestyle and set a games record in winning the 100 m freestyle and a world record in winning the 50 m freestyle. In the 2004 games he finished third in the 100m butterfly, won the 100m breaststroke in a dead heat and sop sharing the medal with Germany's Daniel Clausner, he also retained the gold in the 50m freestyle and 100m freestyle where he broke the games record. At the 2008 games he finished seventh in the 100m freestyle and won bronze in both the 100m butterfly and 50m freestyle where he swam quicker than the games record had been before the heats. The 2000 Paralympic Games were held in Sydney, Australia, from 18 to 29 October. Andrey has competed in three Paralympics in 2000, 2004 and 2008 winning multiple medals each time.\\
\textit{target} &\textbf{In 1993 Sydney host}\\
\textit{output} & \leavevmode\color{blue} Andrey Strokin is a three-time winner of the Sydney Games and host of the annual S13 event.\\
\hline
\textit{wiki-A} & Calcium looping\\
\textit{wiki-B} & Carbon sequestration\\
\textit{context} & Calcium looping (CaL), or the regenerative calcium cycle (RCC), is a second-generation carbon capture technology. It is the most developed form of carbonate looping, where a metal (M) is reversibly reacted between its carbonate form (MCO3) and its oxide form (MO) to separate carbon dioxide from other gases coming from either power generation or an industrial plant. In the calcium looping process, the two species are calcium carbonate (CaCO3) and calcium oxide (CaO). Calcium oxide is often referred to as the sorbent. Carbon sequestration is the process involved in carbon capture and the long-term storage of atmospheric carbon dioxide. The captured carbon dioxide can then be transported to a storage site, used in enhanced oil recovery or used as a chemical feedstock.
\\
\textit{target} & \textbf{storage dioxide other mitigate global}
\\
\textit{output} & \leavevmode\color{blue} Calcium looping allows storage of carbon dioxide and other carbon-rich materials in a carbon-based energy system to mitigate global warming.\\
\hline
\textit{wiki-A} & Quarterback kneel\\
\textit{wiki-B} & Down (gridiron football)\\
\textit{context} & In American football, a quarterback kneel, also called taking a knee, genuflect offense, or victory formation occurs when the quarterback immediately kneels to the ground, ending the play on contact, after receiving the snap. It is primarily used to run the clock down, either at the end of the first half or the game itself, in order to preserve a lead or a win. Especially when the outcome of the game has been well decided, defenses will often give little resistance to the play as a matter of sportsmanship as well as to reduce injury risk on what is a relatively simple play. The quarterback is generally not touched and the act of intentionally taking the knee results in the play being over in all variations of the sport. A down is a period in which a play transpires in American and Canadian football. Although it generally results in a loss of a yard and uses up a down, it minimizes the risk of a fumble, which would give the other team a chance at recovering the ball.\\
\textit{target} & \textbf{distinguishing compared other but rugby}\\
\textit{output} & \leavevmode\color{blue} Quarterback kneel is a distinguishing action compared to other actions but not a form of rugby.\\
\hline
\pagebreak
\hline
\textit{wiki-A} & Sporidiobolus salmonicolor\\
\textit{wiki-B} &  Biosafety level\\
\textit{context} & Sporoidiobolus salmonicolor is a yeast-like fungus in the Basidiomycota more commonly known by the name of its asexual yeast-like state, Sporobolomyces salmonicolor. It has also been reported in AIDS-related infections. The fungus exists predominantly in the asexual state as a unicellular, haploid yeast yet this species can sometimes produce a sexual state when conjugation of compatible yeast cells occurs. The asexual form consists of a characteristic, pink, ballistosporic yeast. A biosafety level is a set of biocontainment precautions required to isolate dangerous biological agents in an enclosed laboratory facility. It is generally considered a Biosafety Risk Group 1 fungus; however isolates of S. salmonicolor have been recovered from cerebrospinal fluid, infected skin, a nasal polyp, lymphadenitis and a case of endophthalmitis.\\
\textit{target} & \textbf{levels range from biosafety highest}\\
\textit{output} & \leavevmode\color{blue} Sporidiobolus salmonicolor is a member of a broad spectrum of fungi from the biosafety level, and is the highest-risk species in its genus.\\
\hline
\textit{wiki-A} & Stanisław Murzynowski\\
\textit{wiki-B} &  Poland\\
\textit{context} & Stanisław Murzynowski (born 1527/8 in the village of Suszyce, died 1553 in Königsberg (Królewiec, today Kaliningrad)) was a Polish writer, translator and a Lutheran activist during the Protestant Reformation. Murzynowski studied at a gymnasium in Królewiec, where he learned Latin, Greek and Hebrew, and then in Wittenberg, where he met Martin Luther and Philipp Melanchthon. He also traveled to Italy in 1547. In 1549 he came back to Królewiec, the capital of Duchy of Prussia which at the time was a secular fief of the Kingdom of Poland. Poland (Polska \textipa{[ 'pOlska ]}), officially the Republic of Poland ([ Rzeczpospolita Polska, links = no ],), is a parliamentary republic in Central Europe. Murzynowski came from a Polish noble family of the Ogończyk coat of arms.\\
\textit{target} & \textbf{state divided administrative subdivisions covering with mostly}\\
\textit{output} & \leavevmode\color{blue} Stanisław Murzynowski became a member of the state council and divided administrative subdivisions of Poland, covering most of the territory with mostly rural population.\\
\hline
\hline
\caption{Examples of generated sentences for given context and target words.}
\label{tab:sample2}
\end{longtable}